\documentclass{article}

\usepackage{natbib}

\usepackage[utf8]{inputenc} 
\usepackage[T1]{fontenc}    
\usepackage{hyperref}       
\usepackage{url}            
\usepackage{booktabs}       
\usepackage{amsfonts}       
\usepackage{nicefrac}       
\usepackage{microtype}      
\usepackage{cancel}
\usepackage{sidecap}
\usepackage{wrapfig}
\usepackage{multirow}
\usepackage{graphicx}
\usepackage{subfigure}
\usepackage{dcolumn}
\usepackage{amsmath, amssymb, mathtools, bm, amsthm}
\usepackage{amsfonts}
\usepackage{gensymb}
\usepackage{placeins}
\usepackage{booktabs} 
\usepackage{todonotes}
\usepackage{hyperref}
\usepackage[ruled,vlined]{algorithm2e}
\usepackage{algorithmic}
\usepackage{bm}
\usepackage{cleveref}

\newtheorem{proposition}{Proposition}

\usepackage{pifont}

\title{Deriving Differential Target Propagation from 
Iterating Approximate Inverses}

\author{%
  Yoshua Bengio\thanks{Mila \& U. Montreal, CIFAR}
}

\begin{document}


\maketitle

\begin{abstract}
We show that a particular form of target propagation, i.e., relying on learned inverses of each layer, which is differential, i.e., where the target is a small perturbation of the forward propagation, gives rise to an update rule which corresponds to an approximate Gauss-Newton gradient-based optimization, without requiring the manipulation or inversion of large matrices. What is interesting is that this is more biologically plausible than back-propagation yet may turn out to implicitly provide a stronger optimization procedure. Extending difference target propagation, we consider
several iterative calculations based on local auto-encoders
at each layer in order to achieve more precise inversions for
more accurate target propagation and we show that these
iterative procedures converge exponentially fast if the
auto-encoding function minus the identity function has
a Lipschitz constant smaller than one, i.e., the auto-encoder
is coarsely succeeding at performing an inversion.
We also propose a way to normalize the changes at each layer
to take into account the relative influence of each
layer on the output, so that larger weight changes
are done on more influential layers, like would happen
in ordinary back-propagation with gradient descent.
\end{abstract}

\section{Introduction and Earlier Work}

In recent years there has been a revival of interest in
exploring analogues of back-propagation which could be more
biologically plausible than back-propagation, for which
concerns were voiced early on~\citep{crick-nature1989}. 
This paper introduces
ideas and mathematical derivations for a family of such
biologically motivated analogues, in line with earlier
work on "target propagation"~\citep{LeCun86,lecun1987phd,lecun1988theoretical,krogh1990cost,Bengio-arxiv2014,Lee-et-al-MLKDB2015-small,lillicrap2020backpropagation}, in which a feedback path provides targets for
each layer, and each layer-to-layer forward
and feedback paths form an auto-encoder, respectively
with its feedforward path encoder and corresponding feedback path decoder.
This paper attempts to provide a solid
theoretical foundation for target propagation
by studying the case where one can use the auto-encoder
to approximately invert the targets of the next layer ($l$) in order
to obtain a target for the previous layer ($l-1$).

See~\citet{lillicrap2020backpropagation} for a review of alternatives
or analogues of backpropagation aimed at bridging the gap between
backpropgation and biology, and in particular of how target propagation
and difference target propagation~\citep{Lee-et-al-MLKDB2015-small} could
potentially be implemented in the brain. The main concerns about a biological implementation
of back-propagation can be summarized as follows:
\begin{enumerate}
    \item The {\em weight transport} or weight symmetry problem
    is that back-propagation requires the backprop
    calculation (going backwards from outputs towards
    inputs) to use the exact same weight matrices (transposed) as those used
    for the forward propagation, but these identical 
    synaptic weights would have to be on different neurons.
    \item The non-linearities used in the forward and backward phase
    of backpropagation are not of the same type and in addition those
    on the backward path need to be numerically coordinated with the corresponding ones in the
    forward path (since they should multiply their input by the first derivative
    of the corresponding forward non-linearities).
    \item The brain has recurrent connections (both feedback
    and lateral ones) which do not disappear while the signals travel
    on the feedforward paths.
    \item Backpropagation needs a clean separation between examples
    and between the forward phase and backward phase, with the forward phase
    activations somehow being held while the backward phase proceeds
    because a combination forward phase and backward phase quantities
    is needed to perform the synaptic weight updates. Various analogues
    of backpropagation such as difference target propagation and the proposal
    made here require a form of time-multiplexing and synchronization in
    order to separate the feedforward and feedback phases, while the brain
    seems to use feedback paths all the time.
    \item Dale's law constrains the signs of the synaptic weights (depending
    on the neuron type), while they are unconstrained in most artificial
    neural network implementations, including those aiming at biological
    plausibility.
    \item Neurons communicate with spikes, not with real values.
    \item Backpropagating through time (BPTT, used to handle recurrent networks)
    requires (a) storing all past states
    of the brain and (b) traversing these states backwards to compute updates
    (c) waiting for a long time before making these updates (to be able to
    capture sufficiently long dependencies). Instead, the brain seems to
    make online updates (or possibly daily updates for longer-term consolidation
    during sleep).
\end{enumerate}
This paper (and target propagation in general) helps with the problem (1)
of weight symmetry or weight transport~\citep{grossberg1987competitive,crick-nature1989,liao2015important},
the problem (2) of having to match non-linearities as well, the problem (3)
of explaining (to some extent) the feedback and lateral recurrence, and it
suggests a way to deal with the problem (4) of synchronization and phases
(using neuromodulating gating, e.g., with acetylcholine). This paper does
not discuss spiking (but rates can approximately be obtained in various ways
by averaging over time and groups of neurons), Dale's law nor BPTT.

Many other approaches have been proposed to deal
    with the weight transport problem, notably feedback alignment~\citep{Lillicrap-et-al-nature2016},
    by showing exact transposes are not always needed (although the approach is still challenged
    by more complex tasks like ImageNet object recognition~\citep{Bartunov-et-al-2018}).
    This is in addition to target propagation approaches~\citep{lecun1987phd,Bengio-arxiv2014,Lee-et-al-MLKDB2015-small,ororbia2019biologically,lillicrap2020backpropagation}, which
    train auto-encoders to obtain feedback weights.
Like many other approaches with
a feedback path, the approach discussed here assumes that the feedback
path mainly influences the computation of targets for the feedforward
path. However, variations proposed here use the recurrent computation
to settle to approximate inverses of the output target at each layer (so that
moving the feedforward activations to these targets would produce the
desired output on the final layer). 
Like Equilibrium Propagation~\citep{Scellier+Bengio-frontiers2017}, this paper
addresses the potential mismatch of non-linearities between the forward
and backward phase, here by appropriately training or iterating over
the feedback paths (whereas Equilibrium Propagation takes advantage of
the propagation of perturbations in the dynamics of a recurrent network
with a static input). Like Equilibrium Propagation, the proposed
iterative procedure for target propagation is a relaxation, but one
that provably converges exponentially fast, with a rate depending
on the reconstruction ability of the auto-encoder (hence the need
to train the feedback weights), and it avoids the weight transport
problem associated with Equilibrium Propagation.
Like ~\citet{podlaski2020biological}, this paper
proposes to use iterative refinements involving the encoder and decoder
at each layer to obtain and then propagate approximate inverses, although
\citet{podlaski2020biological} use fixed random feedback weights whereas
we use a mechanism that requires trained feedback weights
(minimizing a reconstruction error) in order to achieve fast convergence
to a good inverse. Finally, this paper borrows the idea
from GAIT-prop~\citep{ahmad2020gait} that the targets propagated in
target propagation should be only small perturbations of the 
corresponding feedforward activations to facilitate an analysis
of the algorithm based on a linearization of the forward propagation
around the forward phase computation. In comparison to GAIT-prop,
this paper avoids a potentially exponential growth of required learning
rates as one considers layers further from the output, and the proposed
update rule is justified and derived in a different way, although also
starting from the premise of approximate inverses being computable (in our
case via auto-encoder iterations extending the original idea in difference
target propagation).

\section{Layer-wise approximate inverse}

We focus for now on a multi-layer feedforward model and assume that all the layers have the same size, to make the layer-wise inversion feasible and simple. See~\citet{podlaski2020biological} for ways to generalize target-propagation methods to the case of unequal layer sizes. 
See also Section~\ref{sec:skip} to deal with skip connections and architectures which
are not structured as a chain.
Each layer's forward phase activation $h_l$
is computed with its encoder $f_l$:
\begin{align}
    h_l = f_l(h_{l-1})
\end{align}
where $f_l$ is implicitly parametrized by $W_l$, the parameters (weights and biases) used for computing the output of layer $l$, $h_0=x$ is the input vector and $h_L$ is the output layer, with loss $L(h_L,y)$.
The second notation is convenient to describe compositions
of these layer-wise functions, so
\begin{align}
    h_L = (f_L \circ f_{L-1} \circ \ldots f_1)(x).
\end{align}

In addition to learning the feedforward weights $W$,
we consider neural networks with a separate feedback path
with top-down layers $g_l$ and corresponding
feedback parameters $\Omega _l$ to compute an approximate
inverse of $f_l$:
\begin{align}
\label{eq:approx-inverse}
    f_l(g_l(h_{l})) &\approx h_l \nonumber \\
    g_l(f_l(h_{l-1})) &\approx h_{l-1}.
\end{align}
Note that inverses compose, so
\begin{align}
     (f_L \circ f_{L-1} \circ \ldots \circ f_l) \circ (g_l \circ g_{l+1} \circ \ldots g_L) (h_L) \approx h_L
\end{align}
and simple target propagation starting at $\tau_L$ (normally
chosen close to $h_L$) gives rise to 
\begin{align}
    \tau_l = (g_{l+1} \circ \ldots g_L)(\tau_L)
\end{align}
or equivalently, applying a change $\tau_l-h_l$ on layer activations
$h_l$ (to give new values $\tau_l$) would lead to
\begin{align}
\label{eq:recovering-output-delta}
    (f_L \circ f_{L-1} \circ f_{l+1})(\tau_l) \approx \tau_L
\end{align}
which would change $h_L$ in the direction of the gradient
of the loss on the output. Note that the inverse
condition is achieved when we minimize reconstruction
error in the output space (applying the decoder and then
the encoder), which is the opposite of the usual auto-encoder
reconstruction error (where we apply the encoder and then
the decoder). However, the iterative inversion scheme
of Section~\ref{sec:iterative-decoder-output} uses
the regular auto-encoder property of $g_l \circ f_l$
being close to the identity, and the version 
in Section~\ref{sec:biology} minimizes the reconstruction error
of $g_l \circ f_l$ but obtains $\tau_{l-1} \rightarrow f_l^{-1}(\tau_l)$
iteratively.

We will assume that $g$ has enough capacity for approximating
the inverse of $f$. For example it may need an extra hidden
layer or use powerful non-linearities assumed to exist
in dendritic computation in order to obtain the required
flexibility, or the use of iterative refinements to
correct the input or output of the decoder in order 
to approximate the inverse well, as per Sections~\ref{sec:iterative-decoder-input} and~\ref{sec:iterative-decoder-output}. A training objective for enforcing the
above also needs to be defined (such as squared error
on reconstructing pre-activation values). 

\section{Differential Target-Prop when a Single Layer is Changing}
\label{sec:dtp1}

We will first study the case where a single layer $l$ is to have
its weights updated in order to make the output $h_L$ match the target output $\tau_L$.
We will also define differential targets $\tau_l$ for each layer, i.e.,
small perturbations of the forward activations of each
layer obtained by applying and composing
the approximate inverses $g_l$ on the differential output
target $\tau_L$:
\begin{align}
\label{eq:target-prop}
    \tau_L &= h_L - \beta \frac{\partial L(h_L,y)}{\partial h_L}^T \nonumber \\
    \tau_l &= g_{l+1}(\tau_{l+1}) = (g_{l+1}\circ \ldots \circ g_L)(\tau_L)
\end{align}
where $0<\beta\ll 1$ is a small positive constant: small
enough to linearize $g$ and $f$ and large enough to make
computations with limited precision feasible (which possibly 
may require several neurons to represent a single quantity with
enough precision, in the brain).

Because of the availability of an approximate inverse,
we can propagate targets and obtain $\tau_l$ for layer $l$ using Eq.~\ref{eq:target-prop}. It has the property
that if $h_l$ moves to $\tau_l$, we obtain $\tau_L$ as output:
\begin{align}
\label{eq:taul-yields-tauL}
    (f_L \circ f_{L-1} \circ f_{l+1})(\tau_l) & \approx \tau_L \nonumber \\
    L((f_L \circ f_{L-1} \circ f_{l+1})(\tau_l),y) & \approx L(h_L - \beta \frac{\partial L(h_L,y)}{\partial h_L}^T,y) \nonumber \\
    &\approx L(h_L,y) - \beta||\frac{\partial L(h_L,y)}{\partial h_L}||^2 \leq L(h_L,y) 
\end{align}
for $\beta$ small enough (it acts like a learning rate). We can move parameters $W_l$
so that $h_l$ approaches $\tau_l$ by considering $\tau_l$ as a local target
for $h_l$, e.g., using what we now call the single layer update Differential
Target-Prop (DTP1) update, either by moving activations
\begin{align}
    \Delta_{\rm DTP1} h_l = \tau_l-h_l
\end{align}
or with parameter updates (where the update differs from difference target propagation
and other proposals which do not have the normalization factor below):
\begin{align}
\label{eq:delta-dtp}
    \Delta_{\rm DTP1} W_l = \frac{\partial h_l}{\partial W_l}^T \left( \frac{\partial h_l}{\partial W_l} \frac{\partial h_l}{\partial W_l}^T \right)^{-1} (\tau_l-h_l).
\end{align}
The inverse in the above expression is a normalization
factor which goes beyond the usual gradient of the
squared error $(\tau_l - h_l)^2$ with respect to $W_l$,
computed by the rest of the expression. The role of this inverse matrix
is to make sure that when we apply $\Delta_{\rm DTP1} W_l $ to $W_l$, we obtain the change $(\tau_l-h_l)$. Note that
the matrix is normally block-diagonal with a block per neuron,
because the output of one neuron only depends of its incoming synaptic
weights and not (directly) of the synaptic weights of the other neurons.

Now let us see what
the matrix inverse factor provides by computing
the change $\Delta h_l(\Delta_{\rm DTP1} W_l)$ which would result if $\Delta_{\rm DTP1} W_l $ was added to $W_l$. Let us write
$h_l$ as a function of $W_l$ (keeping $h_{l-1}$
fixed):
\begin{align}
    \Delta h_l &= h_l(W_l + \Delta W_l) - h_l(W_l) \nonumber \\
    &\approx \frac{\partial h_l}{\partial W_l} \Delta W_l.
\end{align}
Substituting Eq.~\ref{eq:delta-dtp} in the above, we obtain that the change in $h_l$ due to the DTP1 change in $W_l$ is
\begin{align}
\label{eq:dtp1w-yields-tauL}
    \Delta h_l(\Delta_{\rm DTP1} W_l) &\approx \frac{\partial h_l}{\partial W_l} \frac{\partial h_l}{\partial W_l}^T 
    \left( \frac{\partial h_l}{\partial W_l} \frac{\partial h_l}{\partial W_l}^T \right)^{-1} (\tau_l-h_l) \nonumber \\
    &= (\tau_l-h_l) = \Delta_{\rm DTP1} h_l
\end{align}
which was the objective. In other words, by using the
normalization factor, we can recover the desired target
$\tau_l$ on layer $l$ to make $h_L$ go to $\tau_L$.
Whereas we use an explicit feedback path to propagate
the targets down across layers (such that a change
in $h_l$ leads the desired change in $h_L$), we
use the weight update formula (Eq.~\ref{eq:delta-dtp})
to propagate the target change $\tau_l-h_l$ into
a corresponding target change in $W_l$, such
that applying that parameter change would yield the desired
result on $h_l$, and thus on $h_L$.

For example, with $f_l(h_{l-1})=W\sigma(h_{l-1})$ for
some activation non-linearity $\sigma()$, we would get
the following update (shown below, see Eq.~\ref{eq:delta-w-proof}):
\begin{align}
\label{eq:delta-dtp-regular}
    \Delta_{DTP1} W_l =  (\tau_l - h_l) \frac{\sigma(h_{l-1})^T}{||\sigma(h_{l-1})||^2}
\end{align}
which is similar to the delta rule except for
the normalization of the activation vector of the
previous layer.
Note also that $(\tau_l-h_l)$ is already proportional to $\beta$
and that $\beta$ acts like a global learning rate. One could
however also have local adaptation equivalent to a local
learning rate, which could scale or modify the above equation
(e.g. using a momentum method). In Section~\ref{sec:parallel-updates}
we derive a per-layer learning rate to make the updates proportional
to the influence of $W_l$ on $L$.
Let us now derive
Eq.~\ref{eq:delta-dtp-regular} from Eq.~\ref{eq:delta-dtp}
for the special case $f_l(h_{l-1})=W\sigma(h_{l-1})$,
using the Einstein notation to simplify the tensor calculations:
\begin{align}
\label{eq:delta-w-proof}
    h_{l,i} &= W_{l,i,j} \sigma(h_{l-1,j}) \nonumber \\
    \frac{\partial h_{l,i}}{\partial W_{l,i',j}} &= \delta_{i,i'} \sigma(h_{l-1,j}) \nonumber \\
    \left( \frac{\partial h_{l,i}}{\partial W_{l,i',j}} \frac{\partial h_{l,i''}}{\partial W_{l,i',j}} \right)^{-1} &= \left( \delta_{i,i'} \sigma(h_{l-1,j}) 
              \delta_{i'',i'} \sigma(h_{l-1,j}) \right)^{-1} \nonumber \\
    &= \delta_{i,i''} ||\sigma(h_{l-1})||^{-2} \nonumber \\
    \frac{\partial h_{l,i}}{\partial W_{l,i''',j}}  \left( \frac{\partial h_{l,i}}{\partial W_{l,i',j}} \frac{\partial h_{l,i''}}{\partial W_{l,i',j}} \right)^{-1} (\tau_{l,i''}-h_{l,i''}) &= \delta_{i,i'''} \sigma(h_{l-1,j}) \delta_{i,i''} \frac{(\tau_{l,i''}-h_{l,i''})}{||\sigma(h_{l-1})||^{2}}  \nonumber \\
    &= \delta_{i'',i'''} \frac{(\tau_{l,i''}-h_{l,i''})\sigma(h_{l-1,j})}{||\sigma(h_{l-1})||^{2}} 
\end{align}
where $\delta_{i,j}=1_{i=j}$, and for the weights $W_{l,i,j}$ into unit $i$,
\begin{align}
    \frac{\partial h_{l,i}}{\partial W_{l,i,j}}  \left( \frac{\partial h_{l,i}}{\partial W_{l,i',j}} \frac{\partial h_{l,i''}}{\partial W_{l,i',j}} \right)^{-1} (\tau_{l,i''}-h_{l,i''}) &=
     \frac{(\tau_{l,i}-h_{l,i})\sigma(h_{l-1,j})}{||\sigma(h_{l-1})||^{2}} 
\end{align}
as claimed in Eq.~\ref{eq:delta-dtp-regular}.

The above suggests that if we can move parameters
so that $h_l$ approaches $\tau_l$, we would expect the resulting
network to lead a smaller loss on $(x,y)$. We prove a cleaner and
stronger result below about how this rule, if the
inverses are well approximated, leads to a change
$\tau_L-h_L=-\beta \frac{\partial L(h_L,y)}{\partial h_L}$
on the output layer and how this relates to a regular
gradient step through the implicit inverse of a Gauss-Newton
matrix.

\section{DTP1 Updates Approximate Gauss-Newton}
\label{sec:gauss-newton}

Let us try to analyze the $\Delta_{\rm DTP1}$ learning rule
in the limit of $\beta \rightarrow 0$ which linearizes
the encoder and decoder computations. 

We will start from the per-layer approximate inverse of Eq.~\ref{eq:approx-inverse}
and its consequence Eq.~\ref{eq:recovering-output-delta} on the whole
path from an intermediate layer $h_l$ to the output layer $h_L$.
Let us rewrite Eq.~\ref{eq:recovering-output-delta} as follows:
\begin{align}
    \tau_L \approx (f_L \circ f_{L-1} \circ f_{l+1})(h_l + \Delta_{DTP1} h_l)
\end{align}
where $\Delta_{\rm DTP1} h_l=\tau_l-h_l$.
If we do a first-order Taylor expansion of the above equation
around $h_l$ (considering $\Delta_{DTP1} h_l$ is small when $\beta \rightarrow 0$) we get
\begin{align}
\label{eq:linearized-dtp}
    \tau_L \approx h_L + \left( \prod_{k=L}^{l+1} f'_k(h_{k-1}) \right) (\tau_l-h_l). 
\end{align}
Let us write the shorthand
\begin{align}
    J_l = \prod_{k=L}^{l+1} f'_k(h_{k-1}) = \frac{\partial h_L}{\partial h_l}
\end{align}
for the Jacobian matrix for the mapping from $h_l$ to $h_L$, so
\begin{align}
\label{eq:linearized-dtp2}
    \tau_L - h_L \approx J_l (\tau_l - h_l).
\end{align}
Now let us consider the true gradient of the loss with respect
to $h_l$, scaled by $-\beta$ (which would be the SGD step if we were able
to perform one in the space of hidden activities $h_l$):
\begin{align}
\label{eq:delta-sgd}
    \Delta_{\rm SGD} h_l & = -\beta \frac{\partial L(h_L,y)}{\partial h_l}^T \nonumber \\
     &= - \beta J_l^T \frac{\partial L(h_L,y)}{\partial h_L}^T  \nonumber \\
     &= J_l^T (\tau_L - h_L)  \nonumber \\
     &\approx  J_l^T J_l (\tau_l-h_l) 
\end{align}
where for the second line we use the chain rule, for the third line the definition of
$\tau_L$ in Eq.~\ref{eq:target-prop}, and for the fourth line we replace $\tau_L-h_L$ as per 
Eq.~\ref{eq:linearized-dtp2}, so we have
\begin{align}
    -\beta \frac{\partial L(h_L,y)}{\partial h_l}^T \approx J_l^T J_l (\tau_l-h_l).
\end{align}
Now let us compare
the above with the change in $h_l$ which the DTP1 update (Eq.~\ref{eq:delta-dtp}) is trying to achieve:
\begin{align}
    \Delta_{\rm DTP1} h_l &= (\tau_l-h_l) \nonumber \\
      &\approx (J_l^T J_l)^{-1} \Delta_{\rm SGD} h_l \nonumber \\
      &=  (J_l^T J_l)^{-1} (-\beta \frac{\partial L(h_L,y)}{\partial h_l}^T),
\end{align}
where we make use of Eq.~\ref{eq:delta-sgd}.
Hence what we see is that DTP1 is pushing the weights
so that they move $h_l$ not in the direction of the gradient
$-\frac{\partial L(h_L,y)}{\partial h_l}$ but instead
in the direction corresponding to the {\em Gauss-Newton}
direction for updating $h_l$. What this provides
is a form of invariance with respect to the joint effect of 
the parametrization of the forward path and the representation
$h_l$: we want that the change in the resulting function (i.e., in $h_L$)
be the same if we transform both $h_l$ and the forward path while
keeping the same $h_L$. This is a general property of
target propagation since it focuses on finding the change
(in $h_l$ or in $W_l$) which achieves a target $\tau_L$,
irrespective of the specific parametrization in between.
Indeed, if we were to instead perform
a regular gradient descent step with respect to $h_l$, $h_l$ would move in
the direction $\Delta_{\rm SGD} h_l$ and the outcome
would be a change in $h_L$ which is not aligned
with $\tau_L-h_L$ but instead proportional to
$J_l J_l^T (\tau_L - h_L)$:
\begin{align}
    \Delta_{\rm SGD} h_L (\Delta_{\rm SGD} h_l) &= J_l \Delta_{\rm SGD} h_l \nonumber \\
     &= J_l J_l^T (\tau_L - h_L)
\end{align}
using the third line of Eq.~\ref{eq:delta-sgd}.

Note that the main property enjoyed by Gauss-Newton updates, i.e.,
independence of the parametrization, is also
obtained with the parameter update $\Delta_{\rm DTP1} W_l$ and not just by the pseudo-update
of the activations $\Delta_{\rm DTP1} h_l$, because by construction
the former cause the latter, as shown in Eq.~\ref{eq:dtp1w-yields-tauL}.

\section{Iterative Refinement of an Approximate Inverse
by Updating a Correction on the Input of the Decoder}
\label{sec:iterative-decoder-input}

The main concern one may have about the above derivations
is that any reasonable approximate inverse, especially if it is
learned (e.g., with a form of per-layer auto-encoder objective)
is still going to make errors, and that these errors can
be composed and grow across layers. What if the error made by
the approximate inverse $g_l$ is comparable to the targeted
change in $h_l$? 

We consider two approaches to reduce these potential challenges,
which can be combined. First, in this section, we see how
any approximate inverse can be iteratively improved at an exponential
rate by introducing a gradually improved correction at the
input of the decoder. 
Second, in the next section, we consider how to iteratively obtain
a correction at the output of the decoder, again showing
exponential convergence at a rate which depends on how
good the decoder is at inverting the encoder.

To simplify the notation, let us consider in this and the
next section a simple setting with a forward function $f$ with
input $x$ and approximate inverse $g$, with $f(g(y))\approx y$.
Given a target value $\tau_y$ for the
output of $f$,
we are looking for an input target $\tau_x$ such that
\begin{align}
    f(\tau_x) = \tau_y.
\end{align}

Let us introduce a modified argument to the decoder, $u$, which
may initially be $\tau_y$ but which we will iteratively modify such
that $\tau_x=g(u)$ leads to 
\begin{align}
\label{eq:tau_x_condition}
    \tau_x &= f^{-1}(\tau_y) \nonumber \\
    f(\tau_x)&=\tau_y, 
\end{align}
i.e., 
\begin{align}
\label{eq:u-condition}
    f(g(u))=\tau_y
\end{align}
We update $u$ as follows:
\begin{align}
\label{eq:u-recurrence}
    u \leftarrow u + \tau_y - f(g(u)) = r(u)
\end{align}
where $r$ is the map which transforms $u$ iteratively.
If this iteration converges, it means that we have $\tau_y = f(g(u))$
as per Eq.~\ref{eq:u-condition} in order to obtain 
the desired inverse, i.e., Eq.~\ref{eq:tau_x_condition}.

Let us now analyze this relaxation under the assumption that the auto-encoder
only deviates slightly from a perfect inverse, meaning that $f(g(u))-u$
has a Lipschitz constant $\alpha$ which is smaller 1 (ideally much smaller than 1),
i.e., that its Jacobian is close to the identity,
\begin{align}
    \frac{\partial f(g(u))}{\partial u} = I + {\cal E} \nonumber \\
    \frac{\partial f(g(u))-u}{\partial u} = {\cal E}
\end{align}
with ${\cal E}$ a small matrix, i.e., with its largest eigenvalue $\rho({\cal E})$
smaller than a constant $\alpha$ which is also the Lipschitz constant
of $f(g(u))-u$:
\begin{align}
    \rho({\cal E}) \leq \alpha < 1.
\end{align}
Using the Lipschitz
continuity of $r$ with constant $\alpha$, we obtain
using Eq.~\ref{eq:u-recurrence} (with iterates indexed by $t$)
\begin{align}
   ||u_t-u_{t-1}|| &= ||r(u_{t-1})-r(u_{t-2})|| \nonumber \\
    &< \alpha || u_{t-1}-u_{t-2}||,
\end{align}
from which we can conclude that $u_t$ converges
exponentially fast to 0 in $\alpha^t$, where $\alpha$ is closer
to 0 when we have a better inverse approximation and lower
reconstruction errors. Note that convergence implies a perfect inverse is obtained.
This analysis also tells
us the condition for convergence, i.e., that the
spectral radius of the auto-encoder Jacobian deviates from the
identity at most by an $\alpha<1$. In practice, this condition
is easy to obtain with auto-encoders trained to minimize a form
of reconstruction error. To see this, note that $\alpha=1$
would mean that the reconstruction error is of the same order
of magnitude as the reconstruction itself, whereas $\alpha=0.1$
would roughly correspond to a 10\% error. Although we developed
the iterative updates with an intuition based on the feedback
weights already doing a good job, the above analysis shows
that it only need to do a reasonable job. As soon as reconstruction
error gets to be somewhat minimized, we should be able to achieve
the convergence condition easily. From a biological perspective,
initializing the connectivity pattern to obtain approximate matching
of feedforward paths by corresponding feedback paths might already
achieve a sufficient condition for the iterations to converge even
before training kicks up. Furthermore, minimizing reconstruction error
is something which can be done offline, even in the absence of real
data (e.g. in the womb).

\section{Iterative Refinement of an Approximate Inverse by
Updating a Correction on the Output of the Decoder}
\label{sec:iterative-decoder-output}

Instead of correcting $g$ by changing its input,
we can correct it by changing its output, as discussed here.

\subsection{Single-Step Refinement}

Once we have obtained a pair $(x',y')$ with $y'=f_l(x')$ 
such that
$y' \approx \tau_l$ (e.g., using the procedure in the previous section)
we can make an
extra correction which focuses on the error in the space
of $h_{l-1}$, and in the next section we generalize
this idea to obtain an iterative refinement of the inverse,
making a correction on the output side of the decoder.

To lighten notation, let us again focus on a single layer
with an input $x$, output $y$, encoder $f$, decoder $g$,
and output target $\tau_y$ near $y$. The optimal 
additively adjusted input target $\tau_x$
is such that $f(\tau_x)=\tau_y$.
If $g$ misses the optimal $\tau_x$ by some 
error 
\begin{align}
    e=g(\tau_y)-f^{-1}(\tau_y)
\end{align}
and $g$ is smooth, then we can assume approximately the same
error at a nearby point $y'=f(x')$ close to $y$, i.e.,
\begin{align}
    g(\tau_y) - f^{-1}(\tau_y) \approx g(y') - f^{-1}(y') = g(y')-x' = g(f(x'))-x'.
\end{align}
From this estimated local auto-encoder error vector, 
we can approximately solve for $f^{-1}(\tau_y)$:
\begin{align}
    f^{-1}(\tau_y) &\approx g(\tau_y) + x' - g(y') \nonumber \\
    \tau_x &\leftarrow g(\tau_y)+x'-g(y') 
\end{align}
Note that this works to the extent that $f$ and $g$ are
smooth enough (and for both of them to be smooth, we want
their Jacobians to have singular values close to 1, which
is good for back-propagation as well). If we had obtained
$\tau_x$ by iterating on the decoder input correction
$\delta_t$ as in the last section, then, instead of
estimating the error $g(\tau_y)-f^{-1}(\tau_y)$ at $\tau_y$ using as
a proxy the error at $y$, $g(y)-f^{-1}(y)=f(g(x))-x$, we can use
the pair $(x', y')$ with $y'$ closest to $\tau_y$ we found 
(normally at the last iteration).
Using the notation from the previous sections, this
would give us the following corrected update:
\begin{align}
\label{eq:tau-improved}
    \tau_{l-1} \leftarrow g_l(\tau_l) + g_l(\tau_l+\delta_{l,t})-g_l(f_l(g_l(\tau_l+\delta_{l,t})))
    = g_l(\tau_l) + g_l(u_t)-g_l(f_l(g_l(u_t)))
\end{align}
where $t$ is the last step of the iterative refinement
and $\delta_{l,t}$ the last increment obtained when trying
to invert $f_l$ at $\tau_l$.

\subsection{Iterative Correction of the Decoder Output}

We can iterate the type of correction proposed in the previous
subsection in order to exponentially converge to an inverse,
assuming that the regular auto-encoder minus the identity
is sufficiently contracting:
\begin{align}
\label{eq:iterative-tau-out-update}
    \tau_{l-1} \leftarrow m(\tau_{l-1}) = \tau_{l-1} + g_l(\tau_l) - g_l(f_l(\tau_{l-1}))
\end{align}
where we defined the iterate map $m$ and
we can initialize $\tau_{l-1} \leftarrow g_l(\tau_l)$. At any step,
$\tau_{l-1}$ represents a value of the activations at layer
${l-1}$ which was found to be the
best to make $f_l(\tau_{l-1})$ close to $\tau_l$. 
We could also combine the above formula with the one
in the previous section and initialize
$\tau_l\leftarrow g(\tau_l+\delta_{l,t})$
for the last step $t$ in the iterations of the previous section.

We can analyze the convergence of Eq.~\ref{eq:iterative-tau-out-update}
by using ideas similar to those in the previous section.
The trick is to look for a quantity whose iterates have
a rapidly converging (i.e. strongly contractive) map. We again assume that the auto-encoder
does a good job, but this time we need to consider the $g \circ f$
regular auto-encoder rather than the $f \circ g$ reverse auto-encoder.
We will assume that the Jacobian of $g \circ f$ is near the
identity, or more precisely that $m(u)$ has a Lipschitz
constant $\alpha<1$ (which is also a bound on the
spectral radius of $m'(u)$), noting that $m'$ includes
the identity minus the Jacobian of $g \circ f$, which
only leaves the Jacobian of the reconstruction error:
\begin{align}
    m'(u) = I - (I + {\cal E}) = - {\cal E}
\end{align}
and we correspondingly assume that the Lipschitz constant
of $m$ is no more than $\alpha$. With this
Lipschitz condition and $\alpha<1$, we obtain (after noting
the iterates of $\tau_{l-1}$ at step $t$ as $u_t$) that
\begin{align}
    ||u_{t+1} - u_t|| = ||m(u_t) - m(u_{t-1})|| < \alpha ||u_t-u_{t-1}||.
\end{align}
Hence the series of increments $(u_{t+1}-u_t)$ of $\tau_{l-1}$ converges
and it converges exponentially fast, in $\alpha^t$.
Note that at convergence, $g(f(u_t)) \rightarrow g(\tau_y)$,
which means (if $g$ has an inverse) that $f(u_t) \rightarrow \tau_y$.

This analysis suggests that if we use this iteration
to find a good target for the layer below, we should
make sure to minimize reconstruction error of the
regular auto-encoder $g \circ f$.

\section{DTP for Multiple Layers Updated in Parallel}
\label{sec:parallel-updates}

Because $\beta$ is small,
we can consider small variations of $h_l$ to operate on
a linearized forward propagation computation, and the
combined effects of all the increments $\Delta_{\rm DTP1} h_l=\tau_l-h_l$ applied to each layer 
$l$ on the network output $h_L$ simply add up to yield
a change on the output layer. As we have
seen in Section~\ref{sec:dtp1} if $h_l$ moves to $\tau_l$
this will move $h_L$ to $\tau_L$. We also saw that the
DTP1 weight update would move $W_l$ such that $h_l$ moves to $\tau_l$,
and thus that $h_L$ moves to $\tau_L$. Hence doing a DTP1 weight update
in parallel over all the layers yields a change $L (\tau_L-h_L)$ on the output.
We could simply scale down each layer update by $1/L$ but this would
give equal influence on $h_L$ to all the layers. What should be the
relative impact of each layer on the output (and the loss)?
This is the question we study in this section.

If we were to do ordinary gradient descent, the layers with the
largest gradient would move proportionally to the norm of their gradient:
layers with a small gradient (i.e., less influence on the loss) would move less.
Instead, if we apply DTP1 (even scaled by a constant $1/L$) to all
layers in parallel, the layers with the most influence on the loss
(which have a smaller target change $\tau_l-h_l$ sufficient to
induce the same change $\tau_L-h_L$ in the output) would be changed the least! We do want to keep the
invariance property associated with Gauss-Newton within each layer,
but if we do it separately for each layer we might be doing
something wrong. Also note that {\em any convex combination}
(and not just $1/L$) of the DTP1 updates would produce the
desired target on $h_L$. What is the right weight to put on
each of several changes each of which can individually achieve
the goal? We propose putting more weight on those layers which have more
influence on the loss, just like in gradient descent.

To elucidate this in a simple case,
consider two adjustable input variables $x_1$ and $x_2$
(for now think of them as scalars but later we will think of them as
vectors) and a scalar (loss) function $L(x_1,x_2)$. 
What is the relative magnitude of the loss gradients with respect to each
of them? The gradient with respect to $x_i$ is the
change in the loss $\Delta L(\Delta x_i)$ induced by a change $\Delta x_i$, divided
by $\Delta x_i$, in the limit when the magnitude of $\Delta x_i$
goes to 0, and the magnitude of this ratio seems to be a good measure of the
influence of $x_i$ on $L$.
To be able to compare both limits (for $i=1$ and $i=2$) at
once in terms of their effect on the loss, let us consider a scaled
input change $\frac{\kappa \Delta x_i}{||\Delta x_i||}$ where
we obtain the comparable limits by letting $\kappa$ go to 0 simultaneously
for both changes in $x_1$ and $x_2$. Now we obtain same
size changes in both $x_1$ and $x_2$, so that the magnitude of their relative
gradient (i.e., their relative influence on the loss) can be read off in the corresponding magnitude of the change in loss.
As we show below more formally, the ratio of the influences of the
two components
should be proportional to the ratio of the induced changes in loss
for a same-size change in $x_i$. 

Given some directions of change $\Delta x_i$, one should ask how should we scale them
into $\widehat{\Delta} x_i \propto \Delta x_i$
so that we can update all the $x_i$ in those directions but with the
magnitude of the change, $||\widehat{\Delta} x_i||$ being
proportional to their "influence" on the loss, i.e., a term
we formalize here as a measure of the influence of $x_i$ on $L$, which is $|\frac{\partial L}{\partial x_i}|$ in the scalar case. We thus get
\begin{align}
    \frac{\partial L}{\partial x_i}  &\approx \frac{\Delta L(\Delta x_i)}{\Delta x_i } \nonumber \\
    {\rm influence}(\Delta x_i,L)  &= \frac{||\Delta L(\Delta x_i) ||}{||\Delta x_i ||} \nonumber \\
    {\rm influence}(\Delta x_i,L) &= \frac{1}{\beta}|| \Delta L(\frac{\beta \Delta x_i}{||\Delta x_i||})||
\end{align}
for very small changes (scaled by a small $\beta$).
The influence$(\Delta x_i,L)$ function here measures the magnitude of the effect
of changing component $x_i$ on the loss $L$. 
It is the absolute value of the gradient $\frac{\partial L}{\partial x_i}$
in the case of scalar $x_i$ but otherwise is defined by
the above equations. We need to introduce a $\beta$
to scale $\frac{\Delta x_i}{||\Delta x_i||}$ or otherwise
the change in $x_i$ is too large for the resulting
change in $L$ to indicate only first-order effects.
To compensate we thus need to also divide by $\beta$
outside.

Hence the relative influence of the different
components $x_1$ and $x_2$ on the loss is
\begin{align}
    \frac{{\rm influence}(\Delta x_1,L)}{{\rm influence}(\Delta x_2,L)} &=
    \frac{\Delta L(\frac{\beta \Delta x_1}{||\Delta x_1||})}{\Delta L(\frac{\beta \Delta x_2}{||\Delta x_2||})}.
\end{align}
Now, in our case, $\Delta x_i$ can either be interpreted as a layer change (under DTP1)
or a weight change (also under DTP1). In both cases, the corresponding output change
is the same for all $l$ (Eq.~\ref{eq:taul-yields-tauL} and Eq.~\ref{eq:dtp1w-yields-tauL}),
moving $h_L$ to $\tau_L$. To change $h_l$ or $W_l$
proportionally to their influence on $L$, let us consider
what is that influence:
\begin{align}
    {\rm influence}(\Delta h_l,L) &= \frac{||\Delta L(\Delta_{DTP1} h_l)||}{||\Delta_{DTP1} h_l||} \nonumber \\
    &= \frac{||\frac{\partial L}{\partial h_L} (\tau_L-h_L)||}
            {||\Delta_{DTP1} h_l||} \nonumber \\
    &= \frac{||\frac{\partial L}{\partial h_L} \beta \frac{\partial L}{\partial h_L}^T||}
            {||\Delta_{DTP1} h_l||} \nonumber \\
    &= \frac{\beta ||\frac{\partial L}{\partial h_L}||^2}
            {||\Delta_{DTP1} h_l||}
\end{align}
where we see that the influence is indeed inversely proportional
to $||\Delta_{DTP1} h_l||$ but that it also includes a factor
$||\frac{\partial L}{\partial h_L}||^2$ which is the same
for all layers for a given example $(x,y)$ but which varies
across examples. In the same way that we prefer to make larger
changes on layers with a larger gradient, we prefer to make
larger changes on examples with a larger gradient. Hence,
this analysis is also telling us how to scale the parameter
updates to normalize properly across layers and across examples.

To obtain this, we propose that changes in $h_l$ or $W_l$
be scaled to have as magnitude their influence, times the
global learning rate $\beta$:
\begin{align}
    {\rm update} &= \beta \times {\rm normalized \; update \; direction} \times {\rm influence} \nonumber \\
    \Delta_{DTP} h_l &= \frac{\beta^2 ||\frac{\partial L}{\partial h_L}||^2 }{||\tau_l - h_l||^2} (\tau_l - h_l) \nonumber \\
     &= \frac{||\tau_L-h_L||^2}{||\tau_l-h_l||^2} (\tau_l - h_l) \nonumber \\
     &= \frac{||\tau_L-h_L||^2}{||\tau_l-h_l||^2} \Delta_{DTP1} h_l.
\end{align}
where we have introduced $\Delta_{DTP}$, the differential target-prop update
taking into account the issue of normalizing across layers. Note how the only thing
which is needed to be shared across layers it the norm of the global output loss gradient
$||\tau_L-h_L||^2=\beta^2 ||\frac{\partial L}{\partial h_L}||^2$.

Similarly, we obtain the target parameter change for layer $l$:
\begin{align}
    \Delta_{DTP} W_l &= \frac{||\tau_L-h_L||^2}{||\tau_l-h_l||^2} \Delta_{DTP1} W_l \nonumber \\
     &= \frac{||\tau_L-h_L||^2}{||\tau_l-h_l||^2} \frac{\partial h_l}{\partial W_l}^T \left( \frac{\partial h_l}{\partial W_l} \frac{\partial h_l}{\partial W_l}^T \right)^{-1} (\tau_l-h_l).
\end{align}
Now, when we update all the layers in parallel with DTP, they each contribute in proportion to their influence on the loss.
The special case $f_l(h_{l-1})=W_l \sigma(h_{l-1})$ gives the
update 
\begin{align}
\label{eq:delta-Wl}
    \Delta_{DTP} W_l &= \frac{||\tau_L-h_L||^2}{||\tau_l-h_l||^2}  \frac{(\tau_l-h_l)\sigma(h_{l-1})^T}{||\sigma(h_{l-1})||^2}.
\end{align}

At this point, a natural question is the following: what target should
layer $l$ propagate upstream? The target corresponding to $\Delta_{\rm DTP1} h_l$ which guarantees that lower layers are also trying to
make $h_L$ reach $\tau_L$, or the scaled down target change
$\Delta_{\rm DTP} h_l$ which is weighted to make it proportional
to the influence of $h_l$ on the loss? 
Doing the calculation shows that if we propagate $\Delta_{\rm DTP} h_l$
as a target for lower layers and not account for the scaling
factor $\frac{||\tau_L-h_L||^2}{||\tau_l-h_l||^2}$,
this will result in updates that have this extra factor in them.
It would thus require passing these scaling factors down along
with the feedback path in order to obtain the correct updates
in lower layers. See Section~\ref{sec:numerical} for a related
discussion.

\section{Iterating over Multiple Layers in Parallel}
\label{sec:iterating-multi-layer}

Up to now we have only considered iterating the inverse calculation
on a per-layer basis. This would require waiting for each inverse
to be performed on layer $l$ to obtain a target for layer $l-1$
and start iterations for layer $l-1$. However, this is probably
neither efficient nor biologically plausible. What we propose
instead is the simultaneous optimization of targets at all
layers in parallel. Of course, at least $L$ steps need to be
performed for the information to propagate from $\tau_L$
down to $\tau_1$. Nonetheless, this procedure should converge faster,
since middle layers can start searching for better targets
before the upper layers have converged. This makes sense in
the context that the proposed iterations are rapidly
converging and
gradually provide a more and more appropriate target change
at each layer. Also note that from a biological perspective,
it is plausible that local iterations can be performed quickly
because they involve short local paths. This general
idea is illustrated in Algorithm~\ref{alg:multi-layer-iterations}.

\begin{algorithm}
\caption{Example of parallel multi-layer iterative Differential
Target Propagation and parameter update scheme, for one example $(x,y)$.
The network has $L$ layers, 
each with encoder $f_l$, producing feedforward activation vector $h_l=W_l \sigma(h_{l-1})$ with activation non-linearity $\sigma$. Biases can
be implemented by augmenting the vector $\sigma(x)$ with an additional element with value 1. Each layer also
has a decoder $g_l(u)=\Omega_l \sigma(u)$, used in propagating targets and iteratively
obtain $\tau_{l-1}$ for the layer below, from target $\tau_l$
for layer $l$. It is
advisable to use an invertible non-linearity such as the leaky
ReLU, to facilitate the task of the decoder.
The targets are iteratively updated in parallel
until the lower layer converges to a given StoppingPrecision
hyper-parameter. The overall (input,output) example 
is $(x,y)$ (note that the non-linearity
$\sigma$ is going to be applied to $x$ so some preprocessing may
be appropriate) and there is a loss function $L(h_L,y)$ whose gradient
with respect to $h_L$ is
used to initialize the output target $\tau_L$, with a global learning
rate $\beta \ll 1$. This algorithm shows the weights updated
at the end of the relaxation phase but they can also be
updated on-the-fly with differential equations as per
Section~\ref{sec:online-updates}. The $\forall l$ statements
can be executed in parallel.}
\label{alg:multi-layer-iterations}
\begin{algorithmic}
\STATE // feedforward computation
\STATE $h_0 = x$
\FOR{$l=1$ to $L$}
  \STATE // $f_l$ takes its input from previous layer
  \STATE $h_l \leftarrow f_l(h_{l-1})$ // $=W_l \sigma(h_{l-1})$
  \STATE $n_{l-1} \leftarrow \frac{\sigma(h_{l-1})}{||\sigma(h_{l-1})||^2}$
  \STATE // from now on $f_l$ takes its input from the decoder $g_l$ rather than the previous layer encoder
  \STATE // update decoder weights
  \STATE $\Omega _l \leftarrow \Omega _l + \beta (h_{l-1}-g_l(h_l)) \frac{\sigma(h_l)}{||\sigma(h_l)||^2}$
\ENDFOR
\STATE // initialize the targets
\STATE $\tau_L = h_L - \beta \frac{\partial L(h_L,y)}{\partial h_L}$
\FOR{$l=L$ down to 2}
  \STATE $\tau_{l-1}=g_l(\tau_l)$
\ENDFOR
\STATE // iterative target propagation and improvement
\REPEAT
  \STATE $\forall l,\;\; \tau_{l-1} \leftarrow \tau_{l-1} + g_l(\tau_l) - g_l(f_l(\tau_{l-1}))$
\UNTIL{$||\Delta \tau_1||<$ StoppingPrecision or maximum allowed time elapsed}
\STATE // update feedforward weights
\STATE $\forall l,\;\; W_l \leftarrow W_l +
\frac{||\tau_L-h_L||^2}{||\tau_l-h_l||^2}  (\tau_l-h_l)n_{l-1}^T.$
\end{algorithmic}
\end{algorithm}

\section{Training the Decoders and Architecture Issues}

Up to now we have not discussed an important piece of the puzzle, i.e., the training of the decoder, as well as its architecture. 

\subsection{Expressive Power of the Decoder}

Note that the proposed algorithm requires a separate path with separate neurons for computing the inverses, with these inverses then providing targets for the corresponding feedforward neurons. We have noted that the inverse of a single non-linear layer may require more expressive power than just the same parametrization upside-down. It is thus important that the chosen architecture for the decoder $g_l$ at each level has enough expressive power to approximately invert $f_l$, for any value of the forward parameters $W_l$. For example, the decoder neurons may have non-linearities in its dendrites which enable to compute the
equivalent of an MLP inside a single neuron (i.e., no need to have
an external backprop, but one should then imagine show such
gradients could be computed inside the non-linear dendrites).
On the other hand, if we use one of the iterative procedures
for performing the inverse, the decoder itself may not need
to have the capacity to invert the encoder precisely, so it
could also be that a simple mirror of the feedforward
architecture is sufficient, in that case: there is a
trade-off between the number of iterations needed for
useful inversion and the expressive power of the decoder.

An interesting extension of the proposal in this paper is to
make the decoder have its own hidden layer. This could easily be
obtained as follows, in the context of encoder-input correction.
Imagine we had another group of neurons which predict $u_l$ or the correction $u_l - \tau_l$
as a function of $\tau_l$. One way to think of this extra hidden layer is as 
a way to initialize $u_l$ (given $\tau_l$). Another way to think about it is that
we now have a 2-layer decoder with a training signal for both layers of the decoder.
The advantage of this second decoder layer could be even faster convergence of the inverse,
so it could be an interesting trade-off between computation time and computational resources.
Note also how this principle of iterating over intermediate layers to achieve a global
consistency (here the inversion coherence) might provide a more general way to
obtain hidden layer targets, and similar ideas have been explored in the deep learning literature~\citep{carreira2014distributed,lee2015deeply,taylor2016training,choromanska2019beyond}.

\subsection{Reconstruction Objective}

An important question is whether the decoder weights should be updated to optimize a regular auto-encoder reconstruction error ($g \circ f$ should match the identity) or the reverse auto-encoder reconstruction error ($f \circ g$ should match the identity), or both. Depending on the iterative inverse method chosen, it seems that one or the other of these requirements is more appropriate. The regular auto-encoder error is easier to optimize with respect to the decoder $g_l$ since the target for the decoder is readily available in the form of the corresponding feedforward activation $h_{l-1}$. Each time $f_l$ is computed on $h_l$ (which e.g. happens at least in the feedforward phase), there is an opportunity to compute and optimize this reconstruction error. This is the simplest and most biologically plausible solution so we propose it as the first thing to explore. 

Using the reverse auto-encoder objective is more tricky because it requires doing the equivalent of back-propagating through the encoder to get a target on the decoder. Any of the target-propagation methods discussed here could be used, noting that in that configuration, $g_l$ also acts like an approximate inverse for the encoder $f_l$ and so
we can use the same procedure for target propagation. What is more tricky is to imagine the timing of the different phases of these operations having to overlap with the computations being performed for the overall target propagation.

\subsection{Skip Connections and Multiple Paths to Output}
\label{sec:skip}

What if an inverse is not possible, e.g., because the next layer has more units than the previous one? Similarly, what if the architecture has skip connections (as in ResNets for example)?  In those cases, there is no exact inverse, so the best we can hope is to find a target $\tau_{l-1}$ (still close to $h_{l-1}$) which minimizes the
reconstruction error of the next-level target, when projected
down through $g$. This is already what Eq.~\ref{eq:iterative-tau-out-update} does.

Let us now consider skip connections and more generally
the architectures which are not in the form of a linear chain.
For example, layer $A$ may feed both layers $B$ and $C$ (and they
in turn influence the output layer). The targets $\tau_B$
and $\tau_C$ each individually guarantee that $\tau_L$ can
be obtained by $h_L$, if either $h_A$ or $h_B$ were changed
to their respective target. We have an encoder $f_B$ which maps the inputs
of $B$ (including $A$) to $h_B$, and we have a decoder $g_{BA}$
which maps a target $\tau_B$ into a target $\tau_{BA}$ which would
be sufficient to make $h_B$ changed to $\tau_B$ if $h_A$ was
set to $\tau_{BA}=g_{BA}(\tau_B)$ or to the result of an
iterative process involving $f_B$ and $g_{BA}$. We have a similar
decoder $g_{CA}$ which produces (iteratively or not) a target $\tau_{CA}$ for $h_A$
sufficient to bring $h_C$ to $\tau_C$. Since either picking
$\tau_A=\tau_{AB}$ or $\tau_A=\tau_{CA}$ is sufficient to make
$h_L$ reach $\tau_L$, and since $\beta$ is small enough
to make the forward maps linearized, it means that setting
$\tau_A = \kappa \tau_{BA} + (1-\kappa) \tau_{CA}$ also works
for any $\kappa$ (but we will restrict ourselves to 
convex combinations, i.e., $0\leq \kappa\leq 1$). What would
be a good choice of $\kappa$? This situation should remind us
of the question of how much weight to give on the different
layers (in the linear chain scenario) when changing all the
layers in parallel (Section~\ref{sec:parallel-updates}).

This leads to the following proposal: set $\kappa$
in the above situation to
\begin{align}
    \kappa = \frac{\frac{1}{||\tau_{BA} - h_A||^2}}{\frac{1}{||\tau_{BA} - h_A||^2}+\frac{1}{||\tau_{CA} - h_A||^2}}
\end{align}
so as to give a weight to each branch which is proportional
to its "influence" on the loss.
This principle can be generalized to more than two descendants
by using as convex weights $\propto 1/||\tau_{XA}-h_A||^2$ for
the feedback from descendant $X$, with the proportionality constant chosen so the
sum of the weights is 1.

\subsection{Numerical Stability}
\label{sec:numerical}

One concerning aspect of the inverse (iterative or not)
and of the Gauss-Newton approximation is that they may lead to numerical
instabilities, as we are dividing by quantities (like the
activation layer squared norm) which may be small, pushing
$\tau_l$ far from $h_l$ or making weights change too drastically.
One possibility to avoid such problems
is to bound the maximum deviation of $\tau_l$ from $h_l$
in the computation of $\tau_l$, and to bound the maximum
change in the synaptic weights, but this would possibly
introduce biases. An alternative we favor is to scale
the target changes at each layer to maintain a uniform
magnitude of these changes across layers but keep track of 
the scaling required to make the weight updates conform
to the DTP update. This can be done by propagating both
the target (via the feedback weight) as well as a scalar
quantity, such that their product corresponds to the DTP
target changes. That scalar would then be used as an
local learning rate for each layer in order to recuperate
the correct DTP weight changes.

\section{Biological Implementation}
\label{sec:biology}

We have not discussed much how the ideas in this
paper could be implemented by biological neurons.
See in particular the review from~\citet{lillicrap2020backpropagation} for
a general discussion, especially around target
propagation's  biological implementation.
In terms of architecture, we can assume a similar
structure as in that review paper, which also raises
the question of how to schedule the different phases,
while there is no biological evidence that feedback paths
are inactive during recognition, for example. In fact,
it looks like feedback paths are used not just for
learning but also for forms of attention
and gain control. An interesting possibility
explored here is that instead of having a
sharp distinction between a feedforward phase
and a feedback phase (with different paths becoming
activated), both are always active but with possibly
varying degrees of influence on the feedforward
computation.

\subsection{Updating Weights all the Time}
\label{sec:online-updates}

In terms of biological plausibility, it would make sense
that the proposed synaptic learning rule be applied all
the time and not just at specific moments as is typically
the case in a software implementation of learning algorithms.
On the other hand, in a software implementation where parallel
computation is not completely free, it does make sense
to wait for the right moment to perform the weight updates.
Another desired feature of a good biologically plausible
online updating scheme is that at any point during the settling of propagated targets, it should give good results,
as good as the number of iterations already done allows,
making the algorithm an anytime algorithm, i.e., producing
valid answers whenever it is stopped.

To do this online updating, we need to satisfy two conditions.
First, we want the rule to lead to no update in the feedforward
propagation phase. This is easy to do by letting the output
target be $\tau_L \leftarrow h_L$ (so the
layer-wise targets are automatically converging
to $\tau_l \leftarrow h_l$) during this phase.
Second, while the iterative target computation and target propagation
proceed, there will be intermediate values of $\tau_l$
which are suboptimal. 

Since the normalized input $n_{l-1}=\frac{\sigma(h_{l-1})}{||\sigma(h_{l-1})||^2}$ to layer $l$
does not change while the $\tau_l$ relax, it does not need to
be recomputed in a software implementation and the network can be
recomputing always the same thing in a biologically plausible
implementation where $x$ is fixed during the whole interval
dedicated to processing that example.
Alternatively, if network input
$x$ keeps changing with time, it would be sufficient to 
keep a delayed trace of $n_{l-1}$ in the presynaptic areas, with a delay
equivalent to the time for the combined forward propagation / target
relaxation and propagation. Note that normalization can
be achieved with lateral connections and inhibitory neurons which sum up the squared
activity coming from each layer. The weight updates
can then be performed as follows:
\begin{align}
    \kappa \dot{W}_l = - W_l + \dot{\tau}_l n_{l-1}^T
\end{align}
where $\dot{\tau}_l$ is the error signal computed by the feedback
machinery, with $\tau_l$ initialized with $h_l$ 
and constant during the forward phase (during which there is therefore
no effective update of the weights)
and updated with differential equations according
to the chosen relaxing scheme. As a result,
the total update to $W_l$ is the integral of the above,
which ends up being the last value of $\tau_l$ minus the
initial value of $\tau_l$, times $n_{l-1}^T$. See also
Proposition~\ref{prop:telescoping} for the calculation.

\subsection{CDTP: a Continuously-Running DTP}

To make DTP more biologically plausible, we need to address a number of
issues. One of them is that we would like the same "algorithm" to be
run at each unit of the system, all the time. If some phases or global
coordination are needed, we need to be explicit about them and identify
the signals which need to be communicated across the network to achieve them.

In particular, with DTP as formulated, the same
"hardware", the encoder $f$, has to be used twice but with different
inputs: the feedforward input on one hand, in the feedforward
phase, and the iterated targets from the lower layer on the other
hand, in the feedback phase. There is evidence for gating modulation
in the brain between feedforward and feedback signals which could
be consistent with such a mechanism~\citep{hasselmo2006role}, and we would like
to be more explicit about it here. The results from ~\citet{hasselmo2006role}
suggest that acetylcholine would be involved in the bottom-up vs top-down
modulation as well as correlated with synaptic weight changes and
its cyclic variations linked to the theta rythm oscillations, which
in our interpretation may be linked to the control of feedforward
vs feedback phases. We will denote $\gamma$ for that quantity,
with the convention that $\gamma=1$ corresponds to the feedforward
phase and $\gamma=0$ corresponds to the feedback phase, with $\gamma$
continuously oscillating between 0 and 1 as we go from one "example"
to the next. $\gamma=1$ must either be sustained long enough
for the whole feedforward phase to complete, or more likely from
a biological perspective, $\gamma$ may be different for each layer
and a wave of $\gamma=1$ would travel in the feedforward direction
while the feedforward activation values ($a_l$ below) are computed
and stored.

\subsubsection{Continuous Update Equations}

We propose the following continuous update equations (for now assuming
a global $\gamma$ although a layer-wise travelling wave following the
initial feedforward propagation would also work). We write down the equations
in discrete time but they could as well be written in continuous time.
We will analyze these equations next.\\

\begin{center}
\begin{minipage}{0.8\textwidth}
\begin{align}
\label{eq:hl}
    h_l &\leftarrow f_l(\gamma a_{l-1} + (1-\gamma) g_l(u_l)) \\
\label{eq:al}
    a_l &\leftarrow 1_{\gamma=1} h_l + 1_{\gamma<1} a_l \\
\label{eq:ul}
    u_l &\leftarrow \tau_l + 1_{\gamma<1} (u_l - h_l) = r_l(u_l)\\
\label{eq:taul}
    \tau_{l-1} &\leftarrow \gamma a_{l-1} + (1-\gamma) g_l(u_l) \\
\label{eq:Wl}
    \Delta W_l &= 1_{\alpha<1} \Delta \tau_{l-1} n^T_{l-1} \\
\label{eq:Omegal-1}
    \Delta_1 \Omega_l &= (\tau_{l-1} - g_l(u_l)) \sigma_l(u_l)^T \\
    {\rm or} \nonumber \\
\label{eq:Omegal-2}
    \Delta_2 \Omega_l &= (a_{l-1} - g_l(u_l)) \sigma_l(u_l)^T
\end{align}
\vspace*{-2mm}
\end{minipage}
\end{center}
\vspace*{2mm}

where we have considered two variants of the feedback weights update,
$\Delta \tau_l$ is the temporal variation of $\tau_l$
and $n_l$ is the normalized value of $\sigma(a_l)$ (using lateral
connections to compute the normalization of the layer $l$), i.e.,
\begin{align}
    n_l = \frac{\sigma(a_l) ||a_L-\tau_L||^2}{||\sigma(a_l)||^2\, ||a_l-\tau_l||^2}
\end{align}
and we have chosen the standard linear+nonlinear parametrization
\begin{align}
    f_l(x) &= W_l \sigma(x) \nonumber \\
    g_l(y) &= \Omega_l \sigma(y)
\end{align}
with $\sigma()$ an arbitrary non-linearity (and a bias can be included
by having $\sigma()$ be vector-valued and produce an extra output equal to 1),
as in earlier sections of this paper, although it would be straightforward
to define other parametrizations for the feedforward and/or feedback paths.

\subsubsection{Analysis of the Continuous DTP Equations}

We can verify by inspection of Eqs. \ref{eq:hl} and \ref{eq:al} that $a_l$
acts as a temporary buffer for the activations of the last feedforward
phase (obtained while $\gamma=1$). Conversely, when $\gamma=0$, we
obtain a purely feedback phase where the encoder $f_l$ ignores $a_{l-1}$
and instead is used in an auto-encoder loop which composes $f_l$ and $g_l$
and propagates targets.

\begin{proposition}
\label{prop:final-tau}
If and when the updates of $u_l$ converge, we obtain $\tau_l=h_l$.
\end{proposition}
\begin{proof}
This is trivially proven by solving $r_l(u_l)=u_l$ in Eq.~\ref{eq:ul}.
\end{proof}

\begin{proposition}
\label{prop:inversion}
If and when the updates of $u_l$ converge, we obtain \mbox{$\tau_{l-1} = f^{-1}_l(\tau_l)$}.
\end{proposition}
\begin{proof}
From Proposition~\ref{prop:final-tau}, we obtain that at convergence
\begin{align}
    f^{-1}_l(\tau_l) &= f^{-1}_l(h_l) \nonumber \\
      &= \gamma a_{l-1} + (1-\gamma) g_l(u_l) \nonumber \\
      &= \tau_{l-1}
\end{align}
using Eq.~\ref{eq:hl} for the 2nd line and Eq.~\ref{eq:taul} for the 3rd line
(note we chose Eq.~\ref{eq:taul} precisely to achieve this).
\end{proof}

Let us denote the short-hand $f'_l$, $g'_l$ and $r'_l$ be the Jacobian
matrices corresponding to the functions $f_l$, $g_l$ and $r_l$ respectively
(at the current value of their arguments).
\begin{proposition}
If the spectral radius $\rho(I - (1-\gamma) f'_l g'_l)<1$ 
then the iterates of $u_l$ converge.
\end{proposition}
\begin{proof}
To obtain convergence, note that it is sufficient that the largest eigenvalue of $r'_l$ be less than 1
and differentiate Eq.~\ref{eq:ul} to obtain the proposition statement.
\end{proof}
\begin{proposition}
If the auto-encoder Jacobian $f'_l g'_l = I-\alpha M$ with $\rho(M)=1$
(i.e. the auto-encoder is close to the identity, with spectral radius $\alpha$ for the error)
and $\alpha<1$, then
as $\gamma \rightarrow 0$ (approaching the purely feedback phase) 
we obtain convergence of $u_l$ with updates of $u_l$ converging to 0
at a rate $\alpha^t$ and $\tau_{l-1} \rightarrow g_l(u_l)$.
\end{proposition}
\begin{proof}
With $\gamma$ approaching 0,
\begin{align}
    r'_l &\rightarrow I - f'_l g'_l = I - (I - \alpha M) = \alpha M \nonumber \\
    \rho(r'_l) &\rightarrow \rho( \alpha M) = \alpha.
\end{align}
Hence the Lipschitz coefficient of $r_l$ is bounded by $\alpha$, which means that
iterates of $u_l$ (indexed by $t$ here) approach each other as follows:
\begin{align}
    || u_l(t+1) - u_l(t) || &= || r_l(u_l(t)) - r_l(u_l(t-1)) || < \alpha || u_l(t) - u_l(t-1) || \nonumber \\
    || u_l(t+1) - u_l(t) || & < \alpha || u_l(t) - u_l(t-1) || \nonumber \\
    || u_l(t+1) - u_l(t) || & < \alpha^t || u_l(1) - u_l(0) ||
\end{align}
using the Lipschitz property in the last line.
Finally as $\gamma \rightarrow 0$, from Eq.~\ref{eq:taul} we get $\tau_{l-1} \rightarrow g_l(u_l)$.
\end{proof}

\begin{proposition}
\label{prop:telescoping}
Ignoring the time during which the feedforward activities are computed,
the feedforward weight update (Eq.~\ref{eq:Wl}) is equivalent to the following total
update throughout the combined feedforward-feedback dymamics:
\begin{align}
    \Delta W_l = (\tau_l - a_l) n_{l-1}^T
\end{align}
which is the same as $\Delta_{DTP} W_l$ from Eq.~\ref{eq:delta-Wl}. The initial
updates (while $\gamma=1$) are 0.
\end{proposition}
\begin{proof}
We first note that $\tau_{l-1} \rightarrow a_{l-1}$ at convergence of $u_l$.
The sum of temporal changes in Eq.~\ref{eq:Wl} telescopes because
\begin{align}
    \sum_{t=1}^T \Delta \tau_{l-1}(t) &= (\tau_{l-1}(T) - \cancel{\tau_{l-1}(T-1)}) + 
      (\cancel{\tau_{l-1}(T-1)} - \cancel{\tau_{l-1}(T-2)}) + \ldots \nonumber \\
       & \quad \quad \quad + (\cancel{\tau_{l-1}(1)} - \tau_{l-1}(0)) \nonumber \\
      &= \tau_{l-1}(T) - \tau_{l-1}(0) = \tau_{l-1}(T) - a_{l-1}
\end{align}
The initial updates (while $\gamma=1$) are 0 because $\tau_l$ does not move (Eq.~\ref{eq:taul}),
being fixed to the value $\tau_{l-1}(0)=a_{l-1}$ if the feedforward computation has been performed.
\end{proof}
While the feedforward activations are settling to their value $a_l$ from
their old value (for the previous example), the weight change (if performed)
would aim at predicting the new value of $a_l$
given its previous value, which seems to be a useful feature in a temporal world.
However this could also be hindrance (where there is no temporal order) so in 
Eq.~\ref{eq:Wl} we inserted a $1_{\alpha<1}$ factor to avoid updates until
the feedforward phase as settled (note they would be zero once $a_l$'s have
settled, so long as $\alpha=1$).
This analysis is also compatible with the idea that $\tau_l$ becomes a corrected value
for $a_l$, towards which $a_l$ could move slowly and slightly (not too fast,
to make sure the sequence of $\tau_l$'s from the top $\tau_L$ has time to converge).

\begin{proposition}
The first variant of the feedback weight update $\Delta_1 \Omega_l$ performs an SGD step
with respect to reconstruction error $\frac{1}{2}||a_{l-1} - g_l(f_l(a_{l-1}))||^2$ while $\gamma=1$.
\end{proposition}
\begin{proof}
We see by inspection
that $\Delta_1 \Omega_l$ (Eq.~\ref{eq:Omegal-1}) 
performs an SGD step to minimize $\frac{1}{2}||\tau_{l-1} - g_l(u_l)||^2$.
When $\gamma=1$, Eq.~\ref{eq:taul} gives $\tau_l=a_l$ and $u_l=\tau_l$,
hence $\Delta_1 \Omega_l$ performs an SGD step to minimize
\mbox{$\frac{1}{2}||a_{l-1} - g_l(a_{l})||^2=\frac{1}{2}||a_{l-1} - g_l(f_l(a_{l-1}))||^2$}.
\end{proof}

\begin{proposition}
When $u_l$ has converged, the first variant of the feedback weight update $\Delta_1 \Omega_l$ performs an SGD step
with respect to the inversion error \linebreak[4]
\mbox{$\frac{1}{2}||f_l^{-1}(\tau_l) - g_l(u_l)||^2$} while $\gamma<1$,
i.e., the feedback weight change is trying to hasten the convergence of $g_l(u_l)$ to a value such
that $g_l(u_l)=f_l^{-1}(\tau_l)$. When $\gamma \rightarrow 0$, $\Delta_1 \Omega_l \rightarrow 0$ as well.
\end{proposition}
\begin{proof}
When $u_l$ has converged, $\tau_{l-1} \rightarrow f_l^{-1}(\tau_l)$ (Proposition~\ref{prop:inversion}),
which proves the statement, after noting that the gradient of $\frac{1}{2}||f_l^{-1}(\tau_l) - g_l(u_l)||^2$
gives the $\Delta_1 \Omega_l$ update.
When $\gamma \rightarrow 0$, $\tau_{l-1} \rightarrow g_l(u_l)$, making $\Delta_1 \Omega_l \rightarrow 0$.
\end{proof}

We note that for $\gamma$ varying from 1 to 0, $\tau_{l-1}$ varies linearly from $a_{l-1}$ to $g_l(u_l)$,
making the update interpolate linearly between minimization of the reconstruction error and
helping to hasten the convergence of $g_l(u_l)$ to make $g_l(u_l)$ achieve 
the optimal target propagation $f_l^{-1}(\tau_l)$.

\begin{proposition}
For any $\gamma$ and whether or not $u_l$ has converged,
\begin{align}
    \Delta_1 \Omega_l = \gamma \Delta_2 \Omega_l.
\end{align}
\end{proposition}
\begin{proof}
\begin{align}
    \Delta_1 \Omega_l &= (\tau_{l-1} - g_l(u_l)) \sigma_l(u_l)^T \nonumber \\
      &= (\gamma a_{l-1} + (1-\gamma) g_l(u_l) - g_l(u_l)) \sigma_l(u_l)^T \nonumber \\
      &= (\gamma a_{l-1} - \gamma g_l(u_l)) \sigma_l(u_l)^T \nonumber \\
      &= \gamma \Delta_2 \Omega_l
\end{align}
Using Eq.~\ref{eq:taul} for the 2nd line and the definitions of Eq.~\ref{eq:Omegal-1}
and~\ref{eq:Omegal-2} for the 1st and last line respectively.
\end{proof}
Hence the two updates are equivalent when $\gamma=1$, and whereas
$\Delta_1$ goes to 0 when $\gamma \rightarrow 0$, it is not the case for $\Delta_2$.
The additional pressure $\Omega_l$ receives under $\Delta_2$ can thus best
be understood by considering $\gamma \rightarrow 0$. In that case, the feedback
units are pushed by $\Delta_2$ towards $a_{l-1}$ and not just towards $f^{-1}(\tau_l)$.
This could be seen as a regularizer, especially when there are many inverses:
we would prefer to pick among all the inverses ones which are closer to the feedforward
activations (i.e. require the least changes in the weights). Another  more far-fetched
consideration is that in the case where the loss can be interpreted like an energy
function and the feedback path as a stochastic generative model (if noise is injected
in its computation), then we want that generative path to match the feedforward path,
as in the Wake-Sleep algorithm~\citep{hinton1995wake}.

\section{Conclusion and Future Work}

This is only a proposal, with many uncertain avenues left to explore
and the need to validate these ideas through simulations and confrontation with biological
knowledge (and possibly biological experiments) and machine learning
experiments.
Many variations on the themes introduced here are possible and will need simulations
to disentangle experimentally and understand better.
This proposal opens the door to a style of biologically plausible analogues to backprop which is a variant of target propagation while sharing some properties of error propagation schemes (the targets being
close to the activations)
and can potentially offer optimization advantages compared to SGD with backpropagation, because of the Gauss-Newton approximation implicitly being computed (without having to represent and invert high-dimensional matrices).

One generalization which would be interesting to consider is one in which lower layers
would not just try to minimize the output error but also make the inversion easier for
upper layers, which could be achieved by extending the kind of feedback loop we
analyzed here between consecutive layers to a global feedback look. The general idea
is that we are trying to make all latent layers move away from the current feedforward
activations in a direction which provides less output error but also a better match
between the "prior" top-down information and the "data-driven" bottom-up signal.
An important principle exploited in this paper is that we only want to move slightly
away from the data-driven inference and towards one which is more globally coherent
with other signals (like the target), in order to obtain an update training signal
for all the weights. Like in the Wake-Sleep algorithm, we want to make the generative
part of the model (the feedback loops) slightly more consistent with the data, and
we want to data-driven inference to be slightly more consistent with the global
loss.

An interesting longer-term question is whether this principle
could also be used to implement a limited analogue to backprop through time. 
If the brain has a memory of key past events, as well as predictive computations to link them (by a form of association) both forward and backward in time, then all the ingredients are present to achieve credit assignment through time, in a style similar to Sparse Attentive Backtracking~\citep{Ke-et-al-Backtracking-NIPS2018,kerg2020untangling}.
In the spirit of the Consciousness Prior~\citep{bengio2017consciousness}, these credit assignment
calculations through time 
may need not be for all the units in some circuit: it would be enough
that the forward and backward associations involve only a few
relevant high-level semantic variables. The learned back-through-time
target propagation could use attention mechanisms to choose
aspects of past moments (events) which are relevant and need
to be "fixed" to achieve some current targets.
Attention could also be used in a top-down fashion when
calculating targets, to focus the target change in the most
relevant neurons. This is especially relevant when the
lower layer for which a target is computed has more neurons
than its upper layer.

\bibliographystyle{plainnat}
\bibliography{main}

\end{document}